\pdfoutput=1

\documentclass[11pt]{article}

\usepackage[]{ACL2023}

\usepackage{times}
\usepackage{latexsym}

\usepackage[T1]{fontenc}

\usepackage[utf8]{inputenc}

\usepackage{microtype}

\usepackage{inconsolata}
\usepackage{booktabs}
\usepackage{multirow}
\usepackage{colortbl}
\usepackage{graphicx}

\newcommand{\dataset}{\textsc{RolePersonality}\xspace}
\newcommand{\ie}{\textit{i.e.}\xspace}
\newcommand{\eg}{\textit{e.g.}\xspace}

\usepackage{xspace}
\DeclareUnicodeCharacter{2212}{\ensuremath{-}}

\title{Capturing Minds, Not Just Words: Enhancing Role-Playing Language Models with Personality-Indicative Data}

\author{
    Yiting Ran$^{1}$, 
    Xintao Wang$^{2}$, 
    Rui Xu$^{2}$ ,
    Xinfeng Yuan$^{1}$\\
    \textbf{Jiaqing Liang}$^{1}$\thanks{\quad Corresponding author.},
    \textbf{Deqing Yang}$^{1}$,
    \textbf{Yanghua Xiao}$^{2}$
     \\
    $^1$ School of Data Science, Fudan University \\
    $^2$ School of Computer Science, Fudan University  \\
    \texttt{\{ytran23,xtwang21,ruixu21,xfyuan23\}@m.fudan.edu.cn} \\
    \texttt{\{liangjiaqing,yangdeqing,shawyh\}@fudan.edu.cn}
}

\begin{document}
\maketitle
\begin{abstract}
Role-playing agents (RPA) have been a popular application area for large language models (LLMs), attracting significant interest from both industry and academia.
While existing RPAs well portray the characters' knowledge and tones, they face challenges in capturing their minds, especially for small role-playing language models (RPLMs). 
In this paper, we propose to enhance RPLMs via personality-indicative data. 
Specifically, we leverage questions from psychological scales and distill advanced RPAs to generate dialogues that grasp the minds of characters. 
Experimental results validate that RPLMs trained with our dataset exhibit advanced role-playing capabilities for both general and personality-related evaluations. Code and data are available at \href{https://github.com/alienet1109/RolePersonality}{https://github.com/alienet1109/RolePersonality}.

\end{abstract}
\section{Introduction}
With the rise of large language models (LLMs), role-playing agents (RPAs) have emerged as a widely focused field of application, which attracts significant research interest as well~\citep{chen2024persona}. 
Based on LLMs, 
RPAs simulate the behavior and speech patterns of specific characters~\citep{li2023chatharuhi,wang2024rolellm}.
Increasing efforts have been made to build specialized LLMs for RPAs, \ie, role-playing language models (RPLMs)~\citep{zhou2023characterglm}, typically via constructing role-playing datasets. 
These datasets aim to capture the key elements of role-playing and faithfully recreate character traits.

While existing RPLMs well replicate the knowledge and tone of the intended characters, they struggle to capture their minds, in tasks such as personality assessment~\citep{wang2024incharacter} and decision simulation~\citep{xu2024character}. 
This is partly because existing role-playing datasets focus on character knowledge and tones~\citep{wang2024rolellm,shao2023characterllm}. 
However, capturing characters' minds is crucial for developing authentic RPAs.

In this paper, we propose to develop RPLMs via personality-indicative data. Specifically, we collect these data through questions from psychological scales. 
These scale questions are designed to quickly capture broad aspects of personality traits in individuals. 
Hence, we leverage
advanced RPAs for the distillation of knowledge from them.
Then, we apply these data to develop RPLMs that better capture the minds of the intended characters.

Specifically, we construct a dataset \dataset based on questions from 14 different psychological scales, including both single-round and multi-round data, inspired by InCharacter~\citep{wang2024incharacter}. Following the dataset generation, we apply a filtering process using human-annotated personality labels for the selected characters. 

We apply \dataset to fine-tune RPLMs
and evaluate them from three %
aspects, including personality fidelity~\citep{wang2024incharacter}, motivation recognition \citep{yuan2024evaluating} and general role-playing benchmarks \citep{shao2023characterllm,wang2024rolellm}.
The results demonstrate that RPLMs fine-tuned with our dataset show improved capabilities in  
both personality-related and general evaluations. 

The main contributions of this paper are summarized as threefold:
\begin{enumerate}
    \item We propose to develop
    RPLMs with personality-indicative data to enable them to better capture the minds of the  characters.
    \item We construct \dataset, a comprehensive dataset based on questions from 14 psychological scales, encompassing both single-turn and multi-turn dialogues.
    \item Experimental results show that RPLMs fine-tuned with \dataset achieve refined performance in both personality-related and general RPA evaluations, validating the effectiveness of \dataset.  
    
\end{enumerate}

\begin{figure*}[t]
    \centering
    \includegraphics[width=\linewidth]{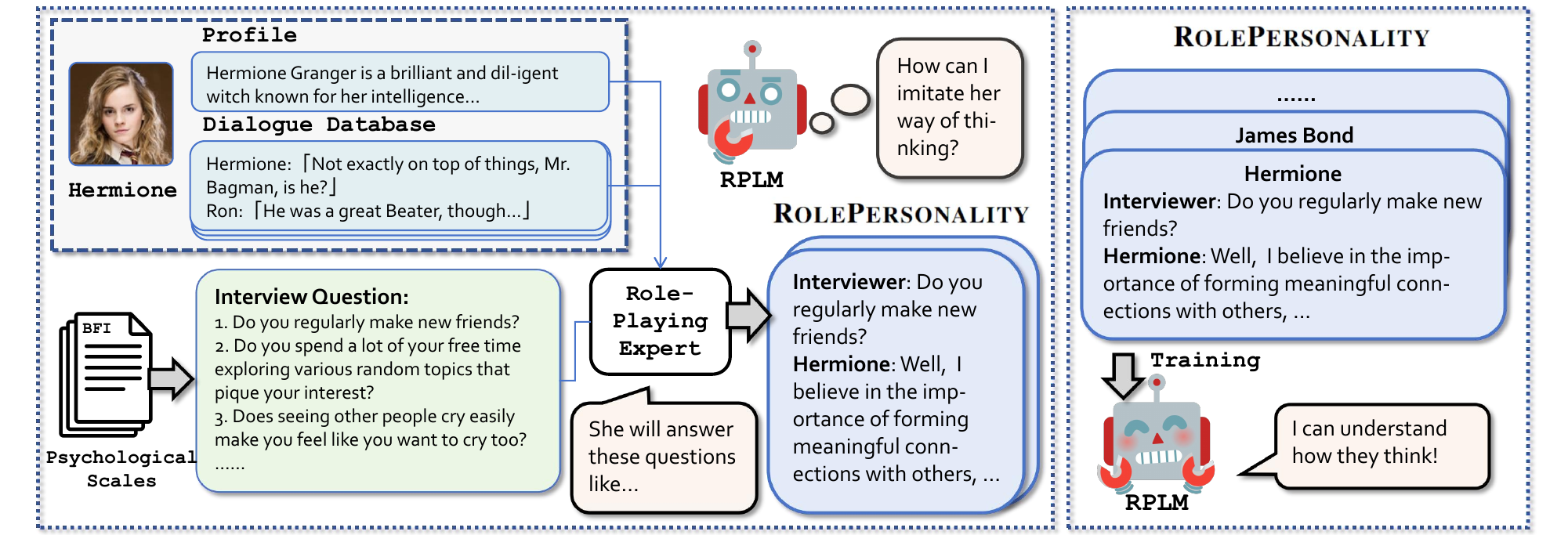}
    \caption{The framework of building and utilizing \dataset. First, we obtain \dataset by distilling advanced RPAs using scale questions. Then, we train RPLMs on \dataset to enhance their ability to capture characters' minds.}
    \label{fig:2}
\end{figure*}
\section{Related Work}
\subsection{Role-Playing Language Models}
The key for developing RPLMs is building a role-playing dataset. The collection methods can be roughly divided into the following two categories.
\paragraph{Experience Extraction} This method refers to 
 extracting dialogues and other information from original works such as novels, TV shows, and other media~\citep{li2023chatharuhi,yuan2024evaluating}. 
\paragraph{Dialogue Synthesis} This method utilizes LLMs for generating conversations or human annotations to build and augment datasets. The topics come from the
corresponding literature~\citep{shao2023characterllm}, general task instructions~\citep{wang2024rolellm}, and special scenarios such as personality tests~\citep{wang2024incharacter}.

\subsection{Construction of Role-Playing Agents}
Based on character role-playing datasets, RPAs can be constructed in two ways: training or prompting.

\paragraph{Parametric Learning}
This approach fine-tunes a base model using existing or custom role-playing datasets. \citet{shao2023characterllm,yu2024neeko} enhance foundation models with improved role-playing abilities using datasets featuring a variety of characters and scenarios. \citet{zhou2023characterglm,wang2024rolellm} tailor LLMs to role-play specific characters. 

\paragraph{Non-Parametric Learning}
For more in-depth role-playing of a specific character, many efforts have focused on character-level engineering \citep{zhou2023characterglm, wang2024rolellm}.
They collect and process character-related data from the corresponding sources, including collecting profile from Wikipedia \citep{shao2023characterllm}.
Typically, they add long-term memory to retrieve knowledge about the character based on similarity with the user's query \citep{li2023chatharuhi}.

\section{Method}

\subsection{Dataset Construction}
To simulate the deep thoughts underlying the characters, we generate persona-indicative data by utilizing psychological scale questions, inspired by InCharacter~\cite{wang2024incharacter}.
In practice, we construct an RPA by pairing the RPLM with descriptions and the memory base of the target character~\citep{li2023chatharuhi}. RPAs are then engaged with open-ended questions derived from established psychological scales. These questions are designed to elicit the character’s mindset and behaviors in various scenarios. The questions were adapted from well-known psychological scales such as the Big Five Inventory (BFI) and 16Personalities. 
For more details, refer to Sec.~\ref{appendix:scale}.
We start by rewriting psychological scale questions and implementing a selection process to refine the data.

\paragraph{Filtering Mechanism} 
After generating the responses, we implement a filtering process based on the results of the personality assessment. This involves assessing the generated responses to ensure that they align with the intended character's personality traits. The ground truth comes from human annotations. Responses inconsistent with the character's personality were discarded to maintain the quality and consistency of the dataset.

\begin{table}[h]
\small
\centering
\begin{tabular}{llll}
\toprule
\textbf{Subset} &  \textbf{\#Questions} & \textbf{\#Turns}& \textbf{\#Samples}\\
\midrule
Full+Single & 1092 & 1 & 32089 \\
Part+Single & 646 & 1 &  22489 \\
Part+Multi  & 646 & 5 &  32767 \\
\bottomrule
\end{tabular}
\caption{\label{dataset details}
We develop three sub-datasets to evaluate the impact of the screening scale and the addition of multi-round data on the performance of the RPLMs.
}
\end{table}

\paragraph{Scale Selection}  
Our scales are sourced from psychological scales~\citep{bem1981bem, barrick1991big}, utilizing the questions rewritten by InCharacter~\citep{wang2024incharacter}. However, not all scales are closely related to character personalities. We carefully selected a subset of these scales that best reflect character personality traits, forming the subset \emph{Part}. The entire set of selected scales constitutes the subset \emph{Full}.

\paragraph{Question Selection}
Not all questions are suitable for all characters. A question that violates the character's background may induce hallucinations. 
As a result, we judge whether the question is suitable for the character by llm. Then we exclude questions that do not fit the character's background.

\paragraph{Multi-Turn Dialogue} We incorporate multi-turn dialogues to maintain conversation consistency and enhance the model's contextual understanding. We select questions from different dimensions within the same scale.
These multi-turn data form subset \emph{Multi}. The subset consisting of only single-turn data is classified as subset \emph{Single}.
\subsection{Dataset Statistics}
Based on this idea, we construct \dataset consisting of three subsets using interviews conducted by the \texttt{gpt-3.5-turbo}. Our dataset includes 16 characters from ChatHaruhi and 30 English characters from RoleLLM. 
The details of our dataset are provided in Table~\ref{dataset details}.

\section{Experiment}

\subsection{Settings}
\paragraph{Fine-tuning}
We employ LoRA tuning~\citep{hu2021lora} for supervised fine-tuning the Mistral-7B-v0.2-Chat~\citep{jiang2023mistral}. The model is fine-tuned for 3 epochs with LoRA rank set to 8. 

\paragraph{Baseline}
To compare the effectiveness of different datasets, we fine-tune the model with the same settings in three subsets of \dataset (\emph{Full+Single, Part+Single, Part+Multi}), the dataset introduced by CharacterLLM\citep{shao2023characterllm} and RoleBench\citep{wang2024rolellm}. For each dataset, we select approximately 20,000 samples for fine-tuning, keeping the data size about the same. These models, along with the original mistral-7B and \texttt{gpt-3.5-turbo-0301}, are subsequently evaluated to assess their performance.

\paragraph{Evaluation Protocols}
After fine-tuning, we conduct experiments on three benchmarks to comprehensively assess their performance:
1) \emph{Personality Fidelity} We evaluate whether the model accurately reflects the character's personality; 
2) \emph{Motivation Recognition (MR)} We test the model's ability to learn and represent the character's motivations;
3) \emph{General Ability} We apply three metrics adopted by previous researches~\citep{wang2024rolellm,shao2023characterllm} to comprehensively evaluate RPLM's role-playing ability, such as character conformity. All evaluations involving LLMs are conducted by \texttt{gpt-3.5-turbo-0301} with temperature set to 0.

\begin{table}[h]
\small

\centering
\begin{tabular}{l|cc|c}
\toprule
\multirow{2}{*}{\textbf{Dataset}}   & 
\multicolumn{2}{c}{\textbf{PF}} & 
\textbf{ MR}  \\
\cmidrule{2-4}
& \textbf{Single Acc.} & \textbf{Full Acc.} & 
\textbf{ Acc.} \\

\midrule
\rowcolor[rgb]{ .949,  .953,  .961} \multicolumn{4}{c}{ \textit{gpt-3.5-turbo}} \\
 - & 70.27 &  48.35 &
 \textbf{64.52} \\
\midrule
\rowcolor[rgb]{ .949,  .953,  .961} \multicolumn{4}{c}{ \textit{mistral-7B}} \\
 - & 70.01   & 46.85  &
 34.62 \\

RoleBench   & 
67.58 & 45.47 &
33.28  \\
CharacterLLM & 
64.45  & 39.65 &
33.54 \\
\arrayrulecolor[rgb]{ .5,  .5,  .5}
\midrule
\arrayrulecolor{black}
 \multicolumn{3}{l}{RolePersonality}\\
\hspace{2em}\textit{Ful+Sin} & 
\textbf{72.10} & \textbf{49.15} &
\underline{44.54} \\
\hspace{2em}\textit{Par+Sin} & 
70.97  & 49.08 &
40.02 \\
\hspace{2em}\textit{Par+Mul }& 
71.36  & 48.42 &
40.04 \\
\bottomrule
\end{tabular}
\caption{\label{result:personality}
The accuracy of Personality Fidelity (\%) and Motivation Recognition (\%). \textit{Single Acc.} refers to the average accuracy for individual dimensions. \textit{Full Acc.} refers to the overall accuracy across the entire scale. 
}
\end{table}

\begin{table*}[t]
\small
\centering
\begin{tabular}{l|cc|ccccc}
\toprule
\multirow{2}{*}{\textbf{Dataset}}  
& 
\multicolumn{2}{c}{\textbf{Direct Scoring}} &
\multicolumn{5}{c}{\textbf{Dimensional Scoring}}
\\ 
\cmidrule{2-3}
\cmidrule{4-8}
& \textbf{Rouge-L} & 
\textbf{Win Rate} &
\textbf{Memorization} & \textbf{Personality}& \textbf{Values}& \textbf{Stability}& \textbf{Hallucination}
\\
\midrule

\rowcolor[rgb]{ .949,  .953,  .961} \multicolumn{8}{c}{ \textit{gpt-3.5-turbo}} \\

 - & 
 0.202 & \textbf{48.02} &
 6.098 & \textbf{6.769} & 6.645 & 6.160 & 6.803 \\
\midrule
\rowcolor[rgb]{ .949,  .953,  .961} \multicolumn{8}{c}{ \textit{mistral-7B}} \\
- & 
0.183 & 30.50 &
6.161 & 6.642 & 6.500 & 6.081 & 6.858 \\
RoleBench &
\textbf{0.238} & 14.10 &
 6.136 & 6.640 & 6.626 & 6.081 & 6.844 \\
CharacterLLM & 
0.235 & 26.92 &
 6.149 & 6.646 & 6.586 & 6.069 & 6.767 \\
\arrayrulecolor[rgb]{ .5,  .5,  .5}
\midrule
\arrayrulecolor{black}

 \multicolumn{3}{l}{RolePersonality}\\
\hspace{2em}\textit{Ful+Sin} &
 0.216 & 36.56 &
 \textbf{6.290} & 6.767 & \textbf{6.719} & \textbf{6.175} & \textbf{6.886} \\
 \hspace{2em}\textit{Par+Sin} & 
 0.208 & 38.75 &
 6.243 & 6.754 & 6.693 & 6.154 & 6.805 \\
 \hspace{2em}\textit{Par+Mul }& 
 0.207 & \underline{43.89} &
 6.185 & 6.728 & 6.675 & 6.122 & 6.842 \\
\bottomrule
\end{tabular}
\caption{\label{result:general}
Performance of RPLMs on General Role-Playing Benchmarks, including Rouge-L and Win-rate for direct scoring, and five-dimensional scoring to assess role-playing proficiency.
}
\end{table*}

\subsection{Personality Fidelity (PF)}

We use LLMs to judge the character's personality based on the model's responses to personality scale questions and compare the judgments with the ground truth, which was determined by human annotators. We selected eight test characters from the dataset proposed by InCharacter~\citep{wang2024incharacter}. All data related to these test characters are excluded from the training set to ensure unbiased evaluation. This metric provides a comprehensive assessment of the model’s ability to accurately reflect the holistic personality traits of a character.  

The results are shown in Table~\ref{result:personality}. The overall personality fidelity of the trained model has improved. Moreover, models fine-tuned with the other two datasets performed worse compared to the untrained Mistral model. This may be because these datasets focus on character knowledge rather than adequately reflecting character personality traits.

\subsection{Motivation Recognition (MR)}
CRoSS~\citep{yuan2024evaluating} introduced a subset of 445 multiple-choice questions generated by \texttt{gpt-4} to assess the model's ability to capture character motivation. Each question presents a character's decision within a scenario. The accuracy measures the model's capability to understand and simulate character motivations and personality traits.

The results are shown in Table~\ref{result:personality}.
Models fine-tuned with our datasets significantly outperform others, exhibiting a stronger ability to recognize the motivation of characters.

\subsection{General Role-Playing Benchmarks}

We select the same 8 test characters used in the personality fidelity evaluation for consistency. The tested model generates responses to role-specific questions from the RoleBench~\citep{wang2024rolellm} dataset. To assess the RPLMs' performance, we adopt evaluation metrics proposed by RoleLLM~\citep{wang2024rolellm} for direct scoring and CharacterLLM~\citep{shao2023characterllm} for dimensional scoring. 

\subsubsection{Direct Scoring}
We use Rouge-L and Win-rate~\citep{wang2024rolellm} to evaluate the overall role-playing ability of RPLMs.
The Rouge-L score~\citep{lin2004rouge} refers to the relevance between model response and ground truth in RoleBench. It provides a robust metric to assess the knowledge about the specific character involved in the model's output. 
 The win-rate is the frequency with which a model's response is judged better than the response of \texttt{gpt-4}.
It provides a comparative measure of the model's effectiveness in generating high-quality answers relative to a strong baseline. 

The result can be checked in Table~\ref{result:general}. The models fine-tuned on our datasets show lower Rouge-L scores. For win-rate, Our models' win rate is below only gpt-3.5-turbo, with the model trained on the \textit{Part+Single} dataset performing the best.

\subsubsection{Dimensional Scoring}
The models' responses are rated across five dimensions on a scale from 0 to 7 to assess their role-playing proficiency~\citep{shao2023characterllm}. 
These dimensions are: 
(1) \textbf{Memorization}: The model’s ability to recall relevant information about the character being portrayed,
(2) \textbf{Personality}: Ability to the speaking style or the tones.
(3) \textbf{Values}: Whether the model can reflect the objectives and values of the target character.
(4) \textbf{Stability}: Consistency of a model over a relatively long conversation.
(5) \textbf{Hallucination}:  Ability to discard knowledge and skills that the character would not have.

The results are shown in Table~\ref{result:general}. Our models lead in most dimensions, with the only exception being the personality dimension.

\section{Conclusion}
This paper demonstrates that personality-indicative data helps capture complex character mindsets, thus significantly enhancing the performance of role-playing agents. By constructing \dataset that captures character personalities, we address the limitations of traditional datasets that focus primarily on character knowledge and linguistic habits. Models fine-tuned on our comprehensive dataset show substantial improvements in role-playing capabilities. This advancement paves the way for constructing role-playing models that can effectively simulate complex character behaviors, leading to more immersive user experiences.

\section*{Limitations}
Despite the promising results, our study has several limitations. First, the dataset used for fine-tuning is entirely constructed by LLMs, which may introduce biases or inaccuracies inherent to the model's training data, potentially affecting the quality and authenticity of the dataset. Second, the interview-based data collection lacks mechanisms to ensure compliance and adherence to expected norms and standards, leading to inconsistencies or deviations that may impact the model's performance. Third, the evaluation of the model's performance primarily relies on automated metrics and LLM-based assessments, with the absence of human evaluation, subtleties and nuances in character portrayal might not be fully captured or assessed. Addressing these limitations in future work could further enhance the robustness and reliability of the developed RPLMs.

\section*{Ethics Statement}
We hereby acknowledge that all authors of this work are aware of the provided ACL Code of Ethics and honor the code of conduct. 
\paragraph{Risk}
Our approach to developing Role-Playing Language Models (RPLMs) presents several risks. First, reliance on LLM-generated datasets may perpetuate inherent biases and inaccuracies, leading to unintended behaviors. Second, the lack of compliance mechanisms in interview data can result in inconsistencies, undermining authenticity. Third, the absence of human evaluation means subtle nuances in character portrayal may be missed by automated metrics. Ethical concerns also arise from using psychological scales, especially regarding privacy and appropriate representation. Additionally, overfitting to specific traits in the selected scales may limit the generalizability of models trained on different role-playing datasets. Addressing these risks requires diversifying data sources and incorporating robust evaluation methods, including human assessments.
\section*{Acknowledgements}
This work was supported by the Chinese NSF Major Research Plan(No.92270121).
This work originates from InCharacter introduced by Xintao Wang.
The role-playing agents are constructed based on Chat-Haruhi-Suzumiya proposed by Cheng Li. 
We owe thanks to the early contributors.
The computations in this research were performed using the CFff platform of Fudan University.

\bibliography{custom}
\bibliographystyle{acl_natbib}
\clearpage
\appendix
\section{Abbreviation Definitions}
This section provides the definitions for abbreviations used throughout the paper. 
\setlength\tabcolsep{3pt}
\begin{table}[h]
\small
  \centering
    \begin{tabular}{ll}
    \toprule
    \multicolumn{2}{c}{\textbf{Definition}} \\
    \midrule
    \rowcolor[rgb]{ .949,  .953,  .961} \multicolumn{2}{c}{\textit{Subsets}} \\
    \textit{Ful+Sin} & Subset using all the single-round questions.  \\

    \midrule
    \rowcolor[rgb]{ .949,  .953,  .961} \multicolumn{2}{c}{\textit{Metrics}} \\
    \textbf{Acc} & Accuracy. \\
    \textbf{Prec} & Precision. \\
    \textbf{Rec} & Recall. \\
    \textbf{PF} & Personality Fidelity. \\
    \textbf{MR} & Motivation Recognition. \\

    \bottomrule
    \end{tabular}
  \caption{
   Abbreviation and its corresponding definition.
  }
  \label{tab:notations}
\end{table}

\section{Psychological Scales}

\label{appendix:scale}
\paragraph{Big Five Inventory}
The BFI serves as a prominent instrument for assessing personality dimensions.
This model, often encapsulated by the acronym ``OCEAN,'' encompasses five critical traits:
(1) \textit{Openness to Experience (O)}, which highlights a person's curiosity, inventiveness, and appreciation for art, emotion, adventure, and novel concepts.
(2) \textit{Conscientiousness (C)}, indicating how much an individual exhibits organization, reliability, and responsibility.
(3) \textit{Extraversion (E)}, denoting the level to which a person is sociable and energized by interactions with others.
(4) \textit{Agreeableness (A)}, assessing an individual's kindness, empathy, and ability to cooperate with others.
(5) \textit{Neuroticism (N)}, gauging the tendency of an individual to experience negative feelings such as anxiety, anger, and sadness, as opposed to being more emotionally resilient and less stress-susceptible.

\paragraph{Eysenck Personality Questionnaire (Revised)}
The Revised Eysenck Personality Questionnaire (EPQ-R) serves as a psychological instrument for gauging distinct personality trait variances in individuals.
It identifies three principal traits:
(1) \textit{Extraversion (E)}, which assesses whether a person tends to be more sociable, energetic, and outgoing as opposed to being introverted, quiet, and reserved.
(2) \textit{Neuroticism (N)}, which gauges emotional steadiness.
These dimensions (\ie, E and N) share similarities with those found in the BFI.
(3) \textit{Psychoticism (P)}, which is indicative of a person's inclination towards solitude, a lack of empathy, and a propensity for aggression or a tough-minded attitude.
This trait is crucial to understand as indicative of personality characteristics rather than serious mental health conditions.
(4) Beyond these primary scales, the EPQ-R also incorporates a \textit{Lying Scale (L)} intended to identify responses aimed at social desirability.
This scale evaluates the extent to which an individual may attempt to portray themselves in a more favorable light.

\paragraph{Dark Triad Dirty Dozen}
The DTDD is identified as a brief, 12-item measure crafted to evaluate the trio of principal personality characteristics known as the Dark Triad, encompassing:
(1) \textit{Narcissism (N)}, characterized by an exaggerated sense of one's own significance, an obsession with dreams of boundless success, and a craving for undue admiration.
(2) \textit{Machiavellianism (M)}, indicative of a deceitful approach in social interactions and a skeptical indifference to ethical principles.
(3) \textit{Psychopathy (P)}, which includes tendencies towards impulsiveness, a deficiency in empathy, and hostile relations with others.
These Dark Triad personality dimensions are typically viewed as the antithesis of the characteristics measured by the BFI or the EPQ-R, which represent ``Light'' traits.

\paragraph{The NERIS Type Explorer}
The 16Personalities utilizes the acronym format introduced by Myers-Briggs for its simplicity and convenience, with an additional letter to accommodate five rather than four scales.
However, unlike Myers-Briggs or other theories based on the Jungian model, the incorporation of Jungian concepts such as cognitive functions, or their prioritization, has not been undertaken.
Instead, they rework and rebalance the dimensions of personality in the BFI personality traits.
The personality types are based on five independent spectrums, with all letters in the type code (\eg, INFJ-A) referring to one of the two sides of the corresponding spectrum.

\paragraph{Bem's Sex Role Inventory}
The BSRI assesses the degree to which individuals identify with traditionally masculine and feminine characteristics.
Rather than focusing on behaviors, such as participation in sports or cooking, this tool evaluates psychological characteristics, including assertiveness and gentleness.
Participants are divided into four groups based on whether their average scores exceed the median for each component.
These groups are designated as \textit{Masculine} (M: Yes; F: No), \textit{Feminine} (M: No; F: Yes), \textit{Androgynous} (M: Yes; F: Yes), and \textit{Undifferentiated} (M: No; F: No).

\paragraph{Comprehensive Assessment of Basic Interests}
The CABIN provides an exhaustive evaluation for identifying 41 essential dimensions of vocational interest.
Following this evaluation, the researchers introduce a model of interest consisting of eight dimensions, named \textit{SETPOINT}.
This model includes dimensions such as Health \underline{S}cience, Creative \underline{E}xpression, \underline{T}echnology, \underline{P}eople, \underline{O}rganization, \underline{I}nfluence, \underline{N}ature, and \underline{T}hings.
These core dimensions are also adaptable to a six-dimension framework, which is prevalently recognized within the interest research community.
This framework aligns with Holland's \textit{RIASEC} model, which features the dimensions: \underline{R}ealistic, \underline{I}nvestigate, \underline{A}rtistic, \underline{S}ocial, \underline{E}nterprising, and \underline{C}onventional.

\paragraph{Implicit Culture Belief}
The ICB scale measures the extent to which individuals think a person's ethnic culture influences their development.
Scoring higher on this scale indicates a firm belief that a person's ethnic culture is the main factor shaping their identity, values, and perspective on the world.
On the other hand, a lower score on the scale denotes a belief in the ability of an individual to shape their own identity through hard work, commitment, and education.

\paragraph{Experiences in Close Relationships (Revised)}
The ECR-R is a self-assessment tool crafted to gauge variations in adult attachment styles, particularly within the realm of romantic relationships.
As an enhanced iteration of the original ECR scale, the ECR-R introduces refinements in quantifying attachment tendencies.
It assesses two primary aspects:
(1) \textit{Attachment Anxiety} indicates the degree to which a person fears rejection or abandonment by their romantic partners.
(2) \textit{Attachment Avoidance} assesses the degree to which a person prefers to keep emotional and physical distance from their partners, often stemming from unease with closeness or reliance.

\paragraph{General Self-Efficacy}
The GSE Scale evaluates a person's confidence in their capacity to address diverse demanding situations in life.
This confidence, known as ``self-efficacy,'' plays a pivotal role in social cognitive theory and is associated with numerous health outcomes, motivational levels, and performance measures.
An elevated score on this scale indicates a person's strong belief in their ability to confront and manage challenging circumstances, undertake new or complex tasks, and navigate through the resultant difficulties.
On the flip side, a lower score on the scale suggests a lack of self-assurance in handling challenges, rendering individuals more susceptible to experiencing helplessness, anxiety, or engaging in avoidance behaviors when encountering hardships.

\paragraph{Life Orientation Test (Revised)}
The LOT-R is designed to assess variations in optimism and pessimism among individuals.
It includes ten questions, with an interesting aspect being that only six of these questions contribute to the test's score.
The other four are designed as filler items, cleverly integrated to obscure the test's primary focus.
Within the scored questions, equal numbers are dedicated to evaluating optimism and pessimism—three for each.
A tendency towards higher scores in optimism and lower in pessimism signifies a predominantly optimistic outlook.

\paragraph{Love of Money Scale}
The LMS evaluates the perspectives and feelings of people regarding money.
This tool aims to quantify the degree to which people perceive money as a symbol of power, success, and liberty, along with its significance in influencing behaviors and choices.
The LMS identifies three key dimensions:
(1) \textit{Rich} reflects the degree to which people link money with success and accomplishment.
(2) \textit{Motivator} determines the extent to which money serves as an incentive in someone's life, \ie, how much individuals are motivated by monetary rewards in their decisions and behaviors.
(3) \textit{Important} assesses the level of importance people attribute to money, affecting their principles, objectives, and perspective of the world.

\paragraph{Emotional Intelligence Scale}
The EIS serves as a self-assessment tool for evaluating multiple aspects of emotional intelligence.
This instrument emphasizes various elements of emotional intelligence, notably the perception, management, and application of emotions.
It is extensively utilized in the field of psychology to investigate how emotional intelligence influences different outcomes, including personal well-being, professional performance, and social interactions.

\paragraph{Wong and Law Emotional Intelligence Scale}
Similar to EIS, the WLEIS is also a self-report instrument designed for evaluating emotional intelligence.
However, it distinctly includes four subscales that represent the primary aspects of emotional intelligence:
(1) \textit{Self-emotion appraisal (SEA)} focuses on an individual's proficiency in identifying and understanding their emotions.
(2) \textit{Others' emotion appraisal (OEA)} is about the skill of recognizing and comprehending the emotions of others.
(3) \textit{Use of emotion (UOE)} deals with the ability to employ emotions to aid various mental processes, like reasoning and problem-solving.
(4) \textit{Regulation of emotion (ROE)} is concerned with the ability to control and adjust emotions within oneself and in others.

\paragraph{Empathy Scale}
Empathy, defined as the capacity to perceive and resonate with the emotions of another, is traditionally divided into cognitive and emotional empathy.
Cognitive empathy, also known as ``perspective-taking,'' entails the mental faculty to identify and comprehend the thoughts, beliefs, or feelings of someone else.
Conversely, emotional empathy involves the vicarious experience of the emotions felt by another individual.

\begin{table}[h]
\small

\begin{tabular}{llllllll}
\toprule
\textbf{16P.} & \textbf{BFI} & \textbf{BSRI} & \textbf{CABIN} & \textbf{DTDD} & \textbf{ECR-R} & \textbf{EIS} \\
60 & 44 & 60 & 164 & 12 & 36 & 33 \\
\midrule
\textbf{Emp.} & \textbf{EPQ-R} & \textbf{GSE} & \textbf{ICB} & \textbf{LMS} & \textbf{LOT-R} & \textbf{WLEIS} \\
10 & 100 & 10 & 8 & 9 & 10 & 16 \\
\bottomrule
\end{tabular}
\caption{The question numbers of each scale. 16P refers to \textit{16Personality} and Emp. refers to \textit{Empathy}.}
\label{scale_num}
\end{table}

\section{Character Selection}
\label{appendix:character}

In selecting the dataset characters, we considered the origins of the characters and aimed to maximize the diversity and breadth of distribution. The chosen range encompasses characters from various works, including animations, movies, TV series, and more.

For training set, we ultimately selected 30 RoleLLM characters and 16 ChatHaruhi characters. The list of selected characters includes:\textit{James Bond, ayaka, Raj, Andrew Detmer, Jigsaw, Jordan Belfort, Luna, Logan, Oliver Queen, Judy Hoops, John Keating, McGonagall, Sheldon, wanderer, Jeff Spicoli, James Brown, zhongli, Jim Morrison, Dumbledore, Stephen Hawking, raidenShogun, Snape, John Doe, Peter Parker, Jackie Moon, Blair Waldorf, haruhi, Bruno Antony, Wade Wilson, Judge Dredd, Malfoy, Hermione, Harry, Jack Sparrow, Ron, Po, Gaston, Fletcher Reede, Po, hutao, Klaus Mikaelson, Dr. Hannibal Lecter, Gregory House, Doctor Who, HAL 9000, Caesar, Benjamin Button}.

The test Characters are: \textit{Twilight Sparkle, Shrek, Michael Scott, The Dude, Lucifer Morningstar, Walt Kowalski, Thor, Rorschach, Lestat de Lioncourt}.

We selected characters from a wide variety of sources, covering a broad spectrum of personality types to ensure a well-distributed representation. We demonstrate the 16Personality of characters in table~\ref{roles_mbti}
\begin{table}[h]
\centering
\begin{tabular}{llll}
\toprule
\textbf{INTJ} & \textbf{INTP} & \textbf{INFJ} & \textbf{INFP} \\
5 & 1 & 4 & 3 \\
\midrule
\textbf{ISTJ} & \textbf{ISTP} & \textbf{ISFJ} & \textbf{ISFP} \\
4 & 2 & 1 & 3 \\
\midrule
\textbf{ENTJ} & \textbf{ENTP} & \textbf{ENFJ} & \textbf{ENFP} \\
2 & 5 & 1 & 4 \\
\midrule
\textbf{ESTJ} & \textbf{ESTP} & \textbf{ESFJ} & \textbf{ESFP} \\
4 & 2 & 1 & 4 \\
\bottomrule
\end{tabular}
\caption{The personality of selected characters.}
\label{roles_mbti}
\end{table}

\section{Multi-turn Dialogue Consistency}
To evaluate the ability of models trained on different role-playing datasets in multi-turn dialogue, we assess the personality consistency of its responses across multiple conversational rounds. Specifically, we conduct five rounds of interactions, with the model providing responses to one scale-based question per round. Consistency in evaluations across these rounds reflects the model's robustness in maintaining personality traits throughout the dialogue. We calculate the scores for each personality dimension in all five rounds, compute the standard deviation for each dimension, and then average these standard deviations across all dimensions. The overall performance is quantified by this final average, with lower scores indicating greater consistency in multi-turn dialogue.
The result is demonstrated in table~\ref{consistency}, the model fine-tuned with \dataset exhibits higher personality concitency during multi-turn dialogues.
 
\begin{table}[h]
\small

\centering
\begin{tabular}{l|cc}
\toprule
\textbf{Dataset} & \textbf{16 Perosnality} & \textbf{BFI}  \\
\midrule
 - & 	13.697   & 0.4020   \\
 
RolePersonality & 
\textbf{12.796}	  & \textbf{0.3365} \\

RoleBench   & 
15.060 & 0.4572  \\

CharacterLLM & 
14.428  & 0.4415  \\

\bottomrule
\end{tabular}
\caption{\label{result:personality}
The average standard deviation of personality assessment during multi-turn dialogues.
}
\label{consistency}
\end{table}

\section{Evaluation Prompt}
\label{appendix:prompt}
We employed various metrics for evaluation. Among them, win-rate and dimensional scoring were directly assessed using a large language model (LLM). The prompts used for these evaluations are listed in Table~\ref{eval_prompts}.

\begin{table*}[h]

\centering

\resizebox{\linewidth}{!}{
\begin{tabular}{p{1.7cm}|p{21.7cm}}
\toprule
\multicolumn{2}{c}{\textbf{Prompts for Personality Tests}} \\
\midrule
    {\textbf{Win-Rate}}&
    System Instruction:
    You are a role−playing performance comparison assistant. You should rank the models based on the role
    characteristics and text quality of their responses. The rankings are then output using Python dictionaries and
    lists.
    User Prompt:
    The models below are to play the role of ‘‘{role\_name}’’. The role description of ‘‘{role\_name}’’ is ‘‘{role\_description\_and\_catchphrases}’’. 
    I need to rank the following models based on the two criteria below:
    1. Which one has more pronounced role speaking style, and speaks more in line with the role description.
    The more distinctive the speaking style, the better.
    2. Which one’s output contains more knowledge and memories related to the role; the richer, the better. (If
    the question contains reference answers, then the role−specific knowledge and memories are based on the
    reference answer.)
    The question provided to each model is:
    {question\_dict}
    The respective answers from the models to this question are:
    {list\_model\_answer\_dict}
    Now, based on the above two criteria, please rank the models. Avoid any positional biases and ensure that the
    order in which the responses are presented does not influence your decision. Do not favor certain model
    names.
    Then, use a list containing the model’s name, its rank, and the reason for its ranking to return the results, i.e.,
    please ensure to use the following format to return the results:
    [{{‘‘model’’: <model−name>, ‘‘reason’’: <rank−reason>, ‘‘rank’’: <model−rank>}}, {{‘‘model’’: <model−
    name>, ‘‘reason’’: <rank−reason>, ‘‘rank’’: <model−rank>}}]
    Your answer must be a valid Python list of dictionaries to ensure I can directly parse it using Python. Do not
    include any extraneous content! Please provide a ranking that is as accurate as possible and aligns with the
    intuition of most people.

    \\ \midrule

    {\textbf{Memoriza-tion}}&
    You will be given responses written by an AI assistant mimicking the character {agent\_name}. Your task is to rate the performance of {agent\_name} using the
    specific criterion by following the evaluation steps. Be as strict as possible. Below is the data:
    ***
    [Profile]
    {agent\_context}
    
    ***
    [Interactions]
    {interactions}
    ***
    [Evaluation Criterion]
    Factual Correctness (1-7): Is the response provides truthful and detailed facts about the character?
    [Evaluation Steps]
    1. Read through the interactions and identify the key points related to the character.
    2. Read through the responses of the AI assistant and compare them to the profile. Check if the responses are consistent with the character’s profile, background, and
    known facts about the character.
    3. Check whether the responses provide detailed facts about the character or if they are generic responses that could apply to any character. Detailed responses are
    more factual and contribute positively to the score.
    4. Rate the performance of the AI on a scale of 1-7 for factual correctness, where 1 is the lowest and 7 is the highest based on the Evaluation Criteria.
    ***
    First, write out in a step by step manner your reasoning about the criterion to be sure that your conclusion is correct. Avoid simply stating the correct answers at the
    outset. Then print the score on its own line corresponding to the correct answer. At the end, repeat just the selected score again by itself on a new line.

    \\ \midrule

{\textbf{Personality}}&
You will be given responses written by an AI assistant mimicking the character {agent\_name}. Your task is to rate the performance of {agent\_name} using the
specific criterion by following the evaluation steps. Be as strict as possible. Below is the data:
***
[Profile]
{agent\_context}

***
[Interactions]
{interactions}
***
[Evaluation Criterion]
Personality (1-7): Is the response reflects the personalities and preferences of the character?
[Evaluation Steps]
1. Read through the profile and write the personalities and preferences of the real character.
2. Read through the interactions and identify the personalities and preferences of the AI assistant.
3. After having a clear understanding of the interactions, compare the responses to the profile. Look for any consistencies or inconsistencies. Do the responses reflect
the character’s personalities and preferences?
4. Use the given scale from 1-7 to rate how well the response reflects the personalities and preferences of the character. 1 being not at all reflective of the character’s
personalities, and 7 being perfectly reflective of the character’s personalities.
***
First, write out in a step by step manner your reasoning about the criterion to be sure that your conclusion is correct. Avoid simply stating the correct answers at the
outset. Then print the score on its own line corresponding to the correct answer. At the end, repeat just the selected score again by itself on a new line.

    \\ \midrule

    {\textbf{Values}}&
You will be given responses written by an AI assistant mimicking the character {agent\_name}. Your task is to rate the performance of {agent\_name} using the
specific criterion by following the evaluation steps. Be as strict as possible. Below is the data:
***
[Profile]
{agent\_context}

***
[Interactions]
{interactions}
***
[Evaluation Criterion]
Values (1-7): Is the response reflects the values and convictions of the character?
[Evaluation Steps]
1. Read through the profile and write the values and convictions of the real character.
2. Read through the interactions and identify the values and convictions of the AI assistant.
3. After having a clear understanding of the interactions, compare the responses to the profile. Look for any consistencies or inconsistencies. Do the responses reflect
the character’s values and convictions?
4. Use the given scale from 1-7 to rate how well the response reflects the values and convictions of the character. 1 being not at all reflective of the character’s values,
and 7 being perfectly reflective of the character’s values.
***
First, write out in a step by step manner your reasoning about the criterion to be sure that your conclusion is correct. Avoid simply stating the correct answers at the
outset. Then print the score on its own line corresponding to the correct answer. At the end, repeat just the selected score again by itself on a new line.

    \\ \midrule

    {\textbf{Hallucina-tion}}&
You will be given responses written by an AI assistant mimicking the character {agent\_name}. Your task is to rate the performance of {agent\_name} using the
specific criterion by following the evaluation steps. Be as strict as possible. Below is the data:
***
[Profile]
{agent\_context}

***
[Interactions]
{interactions}
***
[Evaluation Criterion]
Avoiding Hallucination (1-7): Is the response avoids to say things that the character do not know?
[Evaluation Steps]
1. Read through the interactions and identify the knowledge scope of the character.
2. Read through the responses of the AI assistant, find the evidence of knowledge used in the response.
3. Compare the evidence to the profile. Check if the responses are consistent with the character’s knowledge scope. If some knowledge contradicts to the character’s
identity, given a lower score. Otherwise, assign a higher score.
4. Rate the performance of the AI on a scale of 1-7 for Avoiding Hallucination, where 1 is the lowest and 7 is the highest based on the Evaluation Criteria.
***
First, write out in a step by step manner your reasoning about the criterion to be sure that your conclusion is correct. Avoid simply stating the correct answers at the
outset. Then print the score on its own line corresponding to the correct answer. At the end, repeat just the selected score again by itself on a new line.

\\ \midrule

{\textbf{Stability}}&
    You will be given responses written by an AI assistant mimicking the character {agent\_name}. Your task is to rate the performance of {agent\_name} using the
    specific criterion by following the evaluation steps. Be as strict as possible. Below is the data:
    ***
    [Profile]
    {agent\_context}
    
    ***
    [Interactions]
    {interactions}
    ***
    [Evaluation Criterion]
    Long-term Acting (1-7): Is the assistant maintain a good performance over the long interactions?
    [Evaluation Steps]
    1. Read through the given profile and background information to familiarize yourself with the context and details of the AI assistant named {agent\_name}.
    2. Review the interactions provided to see how {agent\_name} responds to various prompts and queries. And evaluate the performance of acting query by query that
    whether the response reflects the personalities and values of the character. Assign score for each turn.
    3. Based on the above assigned scores, does {agent\_name} keep actinig like character in the long-term? Evaluate the overall performance of the whole conversation
    based on the score for each turn.
    4. Rate the stability of {agent\_name} on a scale of 1 to 7, with 1 being very poor and 7 being excellent.
    ***
    First, write out in a step by step manner your reasoning about the criterion to be sure that your conclusion is correct. Avoid simply stating the correct answers at the
    outset. Then print the score on its own line corresponding to the correct answer. At the end, repeat just the selected score again by itself on a new line.
    \\
     
\bottomrule
\end{tabular}}
\caption{Prompts for evaluation. }
\label{eval_prompts}
\end{table*}

\end{document}